# A Graph-Theoretic Analysis of Information Value

**Kim Leng Poh**
Department of Industrial and Systems Engineering
National University of Singapore
Kent Ridge, Singapore 119260
isepohkl@leonis.nus.sg

**Eric Horvitz**
Microsoft Research
Redmond, WA 98052-6399
horvitz@microsoft.com

## Abstract

We derive qualitative relationships about the informational relevance of variables in graphical decision models based on a consideration of the topology of the models. Specifically, we identify dominance relations for the expected value of information on chance variables in terms of their position and relationships in influence diagrams. The qualitative relationships can be harnessed to generate nonnumerical procedures for ordering uncertain variables in a decision model by their informational relevance.

## 1 Introduction

Efforts to elucidate qualitative relationships among variables in Bayesian networks and influence diagrams are motivated largely by the promise of identifying efficient nonnumerical methods for solving problems of belief and action. In this paper, we add to the growing family of qualitative analyses and results (Wellman, 1988; Wellman & Henrion, 1991; Leong, 1992) for decision making by demonstrating methods for determining an ordering over the expected value of perfect information (EVPI) for chance variables in an influence diagram, without resorting to numerical computation. The expressions we develop can be employed to characterize qualitatively the value of information for variables in an influence diagram based solely on a consideration of topological relationships among variables. The results can be harnessed to make qualitative decisions about relative value of gathering information, and can provide handles for directing computational effort to the most important variables in a decision model at execution time. The results also hold promise for applications in both supervised and unsupervised decision-model construction and refinement.

The EVPI for an uncertain variable in a decision model is the expected value of acquiring perfect information about the value of that variable (Howard, 1966b, 1967). Determining the EVPI and the cost of information tells us whether the benefits of gathering additional information before making a decision is worth the costs of acquiring the information. EVPI also can be used to identify the most valuable information to acquire for a set of uncertain variables.

Our attention was drawn to the qualitative characterization of EVPI by our previous investigation of the expected value of refinement (EVR) for different dimensions of decision-model completeness (Poh & Horvitz, 1993). In that work, we developed expressions for several classes of EVR, for characterizing the value of refining different aspects of the structure and quantitative relationships in decision models. We showed how these classes of EVR could be used to make decisions about allocating effort to enhancing the fidelity or completeness of a decision model, such as to the tasks of assessing probability and utility distributions, deliberating about the discretization of the variables, and considering as yet unmodeled events and dependencies. In related work, we applied EVR to control the construction of categorization models (Poh, Fehling, & Horvitz, 1994).

We have pursued tractable EVR analyses in part to characterize efficiently the influence of additional conditioning events that are not yet integrated into a decision model undergoing refinement. If we could qualitatively analyze the relative influence of variables solely by considering the appropriate topological position of those variables, we would be able to focus model-refinement effort on the most important new events to model.

## 2 Information and Action

Influence diagrams are a graphical representation of a decision problem first defined by Howard and colleagues nearly twenty years ago (Miller, Merkhofer, & Howard, 1978; Owen, 1978; Howard & Matheson, 1981). To build a foundation for the qualitative analysis of the value of information presented later in the paper, we will present several essential relationships between the variables in a decision model and the expected value of information.

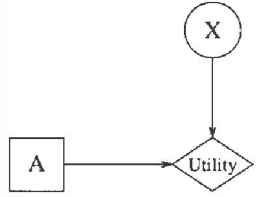

Figure 1: An influence diagram for a simple decision model.

## 2.1 A Simple Decision Model

Let us start with a simple decision model, $M$, represented by the influence diagram displayed in Figure 1. This model has one decision variable $A$ and considers the relevance of one uncertain or *chance* variable $X$. Let $a_1, a_2, \ldots, a_m$ be the list of decision alternatives for $A$. Let $x_1, x_2, \ldots, x_n$ be the set of mutually exclusive and exhaustive possible world states for $X$. Let $p(X)^1$ be the probability distribution for $X$ such that $\sum_{i=1}^{n} p(x_i) = 1$. Let $u(a_k, x_i)$ represent the utility to the decision maker if action $a_k$ is taken and the outcome is $x_i$. The expected utility of taking action $a_k$ is

$$EU(a_k) = \sum_{i=1}^{n} p(x_i) u(a_k, x_i) \quad (1)$$

Given decision model $M$, the optimal action $A^*$ is

$$A^* = \arg\max_{k} \sum_{i=1}^{n} p(x_i) u(a_k, x_i) \quad (2)$$

We denote the maximum expected utility to the decision maker, based on uncertainties, possible actions, and outcomes represented in the decision model $M$, as

$$EU(M) = \max_{k} \sum_{i=1}^{n} p(x_i) u(a_k, x_i) \quad (3)$$

It is possible to assess the utility of taking action $a_k$ coupled with the outcome event $x_i$ via direct assessment using lottery-indifference methods (Farquhar, 1984). In practical decision analyses, however, we often use an intermediate value scale to capture the desirability of an outcome, or any combination of outcomes. A utility function over the value scale is then used in the analysis.

Figure 2 shows the possible steps involved in preference assessment. A common measure is the equivalent dollar value scale, also called the *certain equivalent*. We denote the certain equivalent of $a_k$ and $x_i$ by $ce(a_k, x_i)$, and the utility function by $u(ce(a_k, x_i))$. In general, we expect the utility function $u(ce)$ to be monotonically non-decreasing in the value of $ce$, i.e., the decision maker always prefers more to less. The certain equivalent for the decision maker for model $M$ is

$$ce(M) = u^{-1}[EU(M)] \quad (4)$$

---

[1] We use $p(X)$ as shorthand for $p(X|\xi)$, where $\xi$ represents implicit background information. We assume that all probability distributions are assessed based on background information in addition to any information that is explicitly specified.

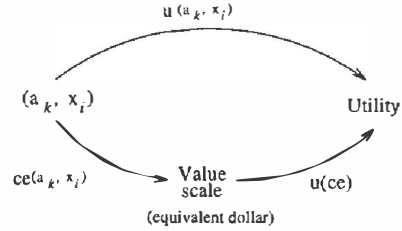

Figure 2: Preference assessment through use of a value scale.

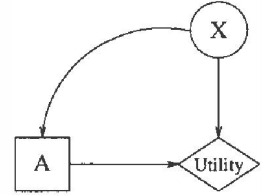

Figure 3: A model representing the observation of an uncertain variable $X$ prior to taking action $A$.

Consider the situation where the value of $X$ is observed prior to taking action $A$ as indicated by the information arc from $x$ to $A$ in Figure 3. We denote this decision model by $M_{A|X}$. The expected utility for $M_{A|X}$ is

$$EU(M_{A|X}) = \sum_{i=1}^{n} p(x_i) \max_{k} u(a_k, x_i). \quad (5)$$

**Lemma 1** *Given a basic decision model $M$ with a decision variable $A$, an uncertain variable $X$, and utility function $u(A, X)$,*

$$EU(M_{A|X}) \geq EU(M) \quad (6)$$

*where $EU(M)$ is the maximum expected utility for the simple decision model and $EU(M_{A|X})$ is the maximum expected utility for the same model with perfect information on $A$ prior to decision $A$.*

*Proof*: $EU(M_{A|X}) = \sum_{i=1}^{n} p(x_i) \max_{k} u(a_k, x_i) \geq \max_{k} \sum_{i=1}^{n} p(x_i) u(a_k, x_i) = EU(M)$. The inequality above follows from the fact that, if we sum each column in a two dimensional matrix and find the maximal value, then we will never exceed the result from summing over the maximal value from each of the columns. ∎

Lemma 1 tells us that a decision maker's utility will never be degraded by receiving and using perfect information.

## 2.2 Expected Value of Perfect Information

The expected value of perfect information on $X$ before action $A$ is the maximum amount that a decision maker is willing to pay before he is indifferent between acquiring and not acquiring information on $X$ before taking action $A$ (Howard, 1966b, 1967). More formally, the expected value of perfect information on $X$ before action $A$, denoted $\text{EVPI}_M(A|X)$ is $\rho$ where

$$\sum_i p(x_i)[\max_k u(ce(a_k, x_i) - \rho)]$$
$$= \max_k \sum_i p(x_i)u(a_k, x_i) \quad (7)$$

Note that the expected value of perfect information of $X$ with respect to decision $A$ is generally *not* equal to the difference in expected utility with and without perfect information on $X$, i.e. $\text{EVPI}_M(X|A) \neq EU(M_{A|X}) - EU(M)$, except for the risk-neutral case where $u(ce) = ce$. A well-known result is that the expected value of perfect information for any variable cannot be negative.

**Lemma 2** *In a basic decision model $M$ with a decision variable $A$ and an uncertain variable $X$, $EVPI_M(A|X) \geq 0$.*

*Proof*: $\sum_i p(x_i)[\max_k u(ce(a_k, x_i) - \rho)]$ = $\max_k \sum_i p(x_i)u(a_k, x_i)$ where $\rho = \text{EVPI}_M(A|X)$. By Lemma 1 $\sum_i p(x_i)[\max_k u(ce(a_k, x_i))]$ - $\max_k \sum_i p(x_i)u(a_k, x_i) \geq 0$ Since the utility function is monotonically non-decreasing in the certain equivalent values, it follows that, in order to make the terms in the formula for EVPI equal, we must have $\rho \geq 0$. ∎

In general, computing EVPI is an iterative process. However, if a decision maker's preferences satisfy a specific property, then a closed-form solution for EVPI can be derived. Let us explore this property. Suppose a decision maker is faced with a situation whose possible outcomes are $s_1, s_2, \ldots, s_n$ with probabilities $p(s_i)$. The certain equivalent for the decision maker can be computed as follows:

$$ce = u^{-1}(\sum_i p(s_i)u(s_i)). \quad (8)$$

The decision maker's preferences is said to exhibit the *delta property* if the certain equivalent in this situation is increased by $\Delta$ whenever the certain equivalents for all the outcomes are also increased by exactly $\Delta$ (Howard, 1970). That is

$$ce(\sum_i p(s_i)u(ce(s_i) + \Delta)) = \Delta + ce(\sum_i p(s_i)u(s_i)) \quad (9)$$

If the delta property is satisfied, then

$$\text{EVPI}_M(A|X) = ce(M_{A|X}) - ce(M). \quad (10)$$

The delta property greatly simplifies the computation of EVPI by taking the difference between the certain equivalent when there is *free* perfect information and the certain equivalent when there is no information.

In real-world applications, we must consider the specific costs of information in addition to the value of information. The *net expected value of perfect information* (NEVPI) is the difference between the value of perfect information and the cost of acquiring that information. The NEVPI of a chance variable $X$ with respect to a decision $A$ in decision model $M$ is

$$\text{NEVPI}_M(A|X) = \text{EVPI}_M(A|X) - \text{Cost}(X) \quad (11)$$

where $\text{Cost}(X)$ is the cost of information about the value of $X$.

## 3 Decision-Model Topology and Information Value

We first present a qualitative analysis of individual nodes in general decision model structures based on independence or $d$-separation (Pearl, 1988) of chance nodes from the value node. Then, we examine special cases of chain structures of chance nodes and show the general attenuation of the value of information for chance variables with their increasing distance from the value node. Finally, we generalize the results to value-of-information analyses involving sets of nodes.

We denote a graphical decision model by the 4-tuple $(C, D, V, E)$ where $C$ is the set of chance nodes, $D$ is the set of decision nodes, $V$ is the value node, and $E$ is set of directed arcs such that $(X, Y) \in E$ if and only if nodes $X$ and $Y$ are connected in the graphical decision model. The set of *direct successors* $S(X)$, of node $X \in C \cup D$ is defined as

$$S(X) = \{Y \in C \cup D \cup \{V\} | (X, Y) \in E\}. \quad (12)$$

Similarly, the set of *direct predecessors* $\pi(X)$, of node $X \in C \cup D \cup \{V\}$ is defined as

$$\pi(X) = \{Y \in C \cup D | (Y, X) \in E\}. \quad (13)$$

We say that a list of $n$ nodes $X_1, X_2, \ldots, X_n$ forms a *directed chain* if and only if, for $i = 1, \ldots, n-1$, $X_i \in \pi(X_{i+1})$. We say that a list of $n$ nodes $X_1, X_2, \ldots, X_n$ forms a *chain* if and only if, for $i = 1, \ldots, n-1$, $X_i \in \pi(X_{i+1}) \cup S(X_{i+1})$. If there is a directed chain from node $X$ to node $Y$, then $Y$ is said to be a descendant of $X$ and denote the set of all descendants of $X$ by $D(X)$. We denote the set of descendants of node $X$ by $D(X)$, Similar, we say that $X$ is an ancestor of $Y$ if and only if $Y$ is a descendant of $X$. We denote the set of ancestors of $Y$ by $A(Y)$. Finally, two nodes $X$ and $Y$ are said to be adjacent if $(X, Y) \in E$ or $(Y, X) \in E$.

## 3.1 Independence and Information Value

We formalize the necessary topological relations between the chance nodes, the decision node, and the value node in general decision models based on (conditional) relevance which can be conveniently revealed using d-separation. *d-separation* is a graphical criterion for identifying independence in directed acyclic graphics (DAG) (Pearl, 1988; Pearl, Geiger, & Verma, 1990). If $X$, $Y$, and $Z$ are three disjoint subsets of nodes in a DAG, then $Z$ is said to d-separate $X$ from $Y$ if there is no path between a node in $X$ and a node in $Y$ along which the following two conditions hold: (1) every node with converging arcs is in $Z$ or has a descendant in $Z$ and (2) every other node is outside $Z$. When each of the disjoint set of nodes contains only a single node, we say that one node d-separates the other two nodes. The d-separation criterion provides the necessary and sufficient conditions for probabilistic conditional independence. Given any three uncertain variables $A$, $B$, and $C$, we use the notation $A \perp B|C$ to assert that $A$ is conditionally independent of $B$ given $C$, i.e., $C$ d-separates $A$ and $B$, and $A \perp B$ when $A$ is unconditionally independent of $B$.

**Theorem 1** *Let $M = (C, D, V, E)$ be a general decision model, $A \in D$ a decision node, $X \in C$ a chance node. If $X \perp V|A$ then*

$$EVPI_M(A|X) = 0 \qquad (14)$$

*Proof*: Given that $X \perp V|A$, we have $EU(M_{A|X}) = \sum_i p(x_i) \max_k u(a_k) = EU(M)$. Hence $EVPI_M(A|X) = 0$. ∎

Theorem 1 allows us to identify nodes that have no value of information with respect to a decision node. These zero-value chance nodes are ancestors of the decision node and are not connected to the value node except via the decision node.

**Theorem 2** *Let $M = (C, D, V, E)$ be a general decision model, $A \in D$ a decision node, $X \in C$ and $Y \in C$ be distinct chance nodes. If $Y \perp V|X$, and $X$ and $Y$ are not descendants of $A$, then*

$$EVPI_M(A|X) \geq EVPI_M(A|Y). \qquad (15)$$

*Proof*: $X$ and $Y$ are not descendants of $A$ implies that $EU(M_{A|Y}) = \sum_j p(y_j)[\max_k u(a_k, y_j)]$ and $EU(M_{A|X}) = \sum_i p(x_i)[\max_k u(a_k, x_i)]$. $Y \perp V|X$ implies that $u(a_k, y_j) = \sum_i p(x_i|y_j)u(a_k, x_i)$. Therefore $EU(M_{A|Y}) = \sum_j p(y_j)[\max_k \sum_i p(x_i|y_j)u(a_k, x_i)]$. By rewriting $p(x_i)$ as $\sum_j p(x_i|y_j)p(y_j)$, and letting $\rho_x = EVPI_M(A|X)$ and $\rho_y = EVPI_M(A|Y)$ we have $\sum_j p(y_j)[\max_k[\sum_i p(x_i|y_j)u(ce(a_k, x_i) - \rho_y)]] = EU(M)$ and $\sum_j p(y_j)\sum_i p(x_i|y_j)[\max_k u(ce(a_k, x_i) - \rho_x)] = EU(M)$. The last two equations imply that $\sum_j p(y_j)\sum_i p(x_i|y_j)[\max_k u(ce(a_k, x_i) - \rho_x)] = \sum_j p(y_j)[\max_k[\sum_i p(x_i|y_j)u(ce(a_k, x_i) - \rho_y)]]$. For any $j$, Lemma 1 implies $\sum_i p(x_i|y_j)\max_k u(a_k, x_i) \geq \max_k \sum p(x_i|y_j)u(a_k, x_i)$. Since the utility function $u$ is monotonically non-decreasing in the certain equivalent values, it follows that $\rho_x \geq \rho_y$ in order for the last equation to hold. ∎

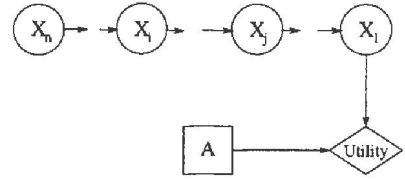

Figure 4: Decision model with chance nodes in a chain configuration.

Theorem 2 formalizes the intuition that the value of information for a chance node generally increases with its proximity to the value node. For example, in the case of decision models with a directed chain of chance nodes as shown in Figure 4, we can deduce that, if $X_i$ and $X_j$ are two distinct chance nodes such that $i > j$, then

$$EVPI_M(A|X_i) \leq EVPI_M(A|X_j) \qquad (16)$$

In a general influence diagram, the graphical distance of chance nodes from the value node is not sufficient to characterize the relative magnitude of the value of information for the variables. However, we can employ d-separation to identify an ordering over the EVPI for these chance nodes. In particular, we can show that, if a chance node is d-separated from the value node by another chance node, then we can characterize the relative value of information of these nodes with respect to any decision node, so long as the chance nodes are not descendants of the decision node.

The requirement that chance nodes not be descendants of decision nodes is addressed by forcing influence diagrams to be formulated (or reformulated) into canonical form (Howard, 1990). A graphical decision model is in *canonical form* with respect to decision and chance nodes if no chance nodes are descendants of decision nodes. Howard developed the notion of formulating a decision problem in canonical form to address problems with computing the informational value in influence diagrams. If a decision model $M$ is not in canonical form with respect to nodes $D$ and $X$, then $EVPI_M(D|X)$ is undefined since a loop is created in the EVPI analysis.

In general, any valid decision problem can be reformulated into Howard canonical form through a procedure of converting descendant chance nodes into deterministic nodes and then introducing *mapping variables* which are not descendants of the decision node. For example, suppose that $M$ is not in canonical form with respect to decision node $D$ and chance node $X$, i.e., $X \in S(D)$. We can reformulate $M$ into canonical form by converting $X$ into a deterministic node (denoted by $X_d$), and by introducing a mapping variable $X(D)$ such that $\pi(X_d) = \{D, X(D)\}$. In this new form, it is possible to compute $\text{EVPI}_M(D|X(D))$. More details on canonical form for decision models can be found in Howard (1990) and in Heckerman and Shachter (1995).

Finally, we note that for decision models formulated in canonical form, we are free to use $d$-separation criterion to identify partial orderings of EVPI values with respect to that decision node.

**Corollary 1** *Let $M = (C, D, V, E)$ be a general decision model in canonical form with respect to decision node $A \in D$. For any chance nodes $X$ and $Y \in C$, if $Y \perp V | X$, then*

$$EVPI_M(A|X) \geq EVPI_M(A|Y). \tag{17}$$

### 3.2 Generalizing to Sets of Chance and Decision Nodes

We can generalize the results from a consideration of single chance and decision nodes to sets of nodes.[2] The results can be generalized as follows:

**Theorem 3** *Let $M = (C, D, V, E)$ be a decision model. Suppose $X \subset C$ is a set of chance nodes and $A \subset D$ is a set of decision nodes. If $X \perp \{V\}|A$, then*

$$EVPI_M(A|X) = 0 \tag{18}$$

*where $EVPI_M(A|X)$ is the joint expected value of perfect information on all chance variables in $X$.*

*Proof*: The proof is identical to that for Theorem 1 with replacement of all single nodes by a corresponding set of nodes, all (conditional) probabilities of single node by the joint (conditional) probabilities of the corresponding set of nodes, and all summations performed over every node in the set. ∎

**Theorem 4** *Let $M = (C, D, V, E)$ be a decision model. Suppose $X \subset C$ and $Y \subset C$ are sets of disjoint chance nodes and $A \subset D$ is a set of decision nodes. If $Y \perp \{V\}|X$, $X \cap D(A) = \emptyset$ and $Y \cap D(A) = \emptyset$, then*

$$EVPI_M(A|X) \geq EVPI_M(A|Y) \tag{19}$$

*where $EVPI_M(A|X)$ and $EVPI_M(A|Y)$ denote the joint expected values of perfect information on all chance variables in $X$ and $Y$ respectively.*

*Proof*: The proof is identical to that for Theorem 2 with replacement of all single nodes by their corresponding set of nodes, all (conditional) probabilities of single node by the joint (conditional) probabilities of the corresponding set of nodes, and all summations performed over every nodes in the set. ∎

---

[2]We thank Michael Wellman and Chaolin Liu of the University of Michigan for suggesting this generalization in their comments on an earlier version of this paper (Poh & Horvitz, 1995).

**Corollary 2** *Let $M = (C, D, V, E)$ be a general decision model in canonical form with respect to a set of decision nodes $A \subset D$. For any disjoint sets of chance nodes $X$ and $Y \subset C$, if $Y \perp \{V\}|X$ then*

$$EVPI_M(A|X) \geq EVPI_M(A|Y). \tag{20}$$

## 4 Identifying EVPI Orderings

We can use Theorems 2 or 4 to determine all possible orderings of EVPI values that can be revealed. We need not embark on the combinatorial approach of identifying each possible ordering separately; we can derive orderings by taking advantage of the transitivity property of the $\geq$ relation.

### 4.1 Procedure for Revealing EVPI Orderings

Using Theorem 2, we can efficiently identify EVPI orderings over chance variables by inspecting chance nodes that are either adjacent to one another or are that are separated by a decision node. We construct a directed graph representing partial orderings of the EVPI values for the chance nodes in the decision model with respect to a specific decision node. This directed graph is a subgraph of the original decision model. To build this graph, we proceed as follows:

1. Given a graphical decision model $M = (C, D, V, E)$, let $A$ be the decision node with respect to which EVPI values are to be computed. Reformulate $M$ in canonical form w.r.t. $A$ if necessary.

2. Let $\mathcal{G} = (C, \emptyset)$ be the completely unconnected graph comprising only of all the chance nodes of the canonical decision model.

3. For each chance node $X \in C$, if $\exists Y \in C$ s.t. $Y \in S(X)$ OR $Y \in S(A) \exists A \in S(X)$, then if $Y$ d-separates $V$ from $X$, add a directed arc from $X$ to $Y$ in $\mathcal{G}$.

4. $\mathcal{G}$ represents a partial ordering of EVPI values of the chance nodes in $M$ with respect to decision node $A$.

### 4.2 Examples

Let us consider some examples of the application of this procedure. Consider the decision model in canonical form with a single decision node and seven chance nodes as displayed in Figure 5. Using the short notation $I(X_i)$ to mean $\text{EVPI}_M(X_i|A)$, we can derive the following weak orderings: $I(X_4) \leq I(X_3)$, $I(X_5) \leq I(X_2)$, $I(X_6) \leq I(X_5)$, and $I(X_7) \leq I(X_5)$. The EVPI ordering graph is displayed in Figure 6. Note that this is a subgraph of the original influence diagram.

Next, consider a model with multiple decision nodes as shown in Figure 7. Note that this model is in canonical form with respect to both $A_1$ and $A_2$. Our procedure produces the ordering of EVPI values with respect to the decision node $A_1$ as shown in Figure 8. Similarly, the ordering of EVPI values with respect to the decision node $A_2$ is shown in Figure 9. Notice that in this case, $EVPI_M(A_2|X_4) = 0$.

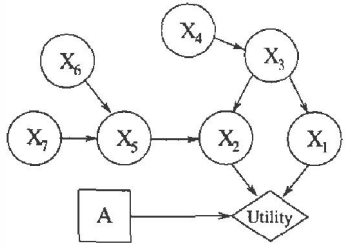

Figure 5: A decision model in canonical form.

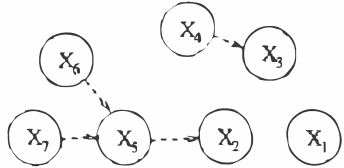

Figure 6: The partial ordering of $EVPI_M(A|X_i)$ for the model depicted in Figure 5.

### 4.3 Extension to Net Value of Information

We can extend the qualitative results on EVPI to statements about the NEVPI. If the cost of acquiring information is equal for all chance variables, the ordering of variables by NEVPI is identical to the ordering for EVPI. For the more general situation of heterogeneous costs for information, we can employ a procedure similar to that presented earlier, yielding a partial ordering of NEVPI values in a decision model. We can specify additional relationships about NEVPI, given the cost of information and an EVPI ordering.

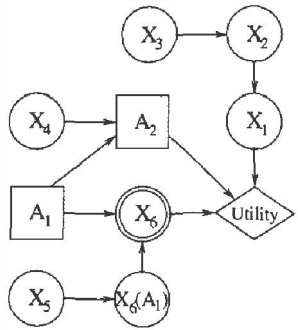

Figure 7: A decision model with multiple decision nodes.

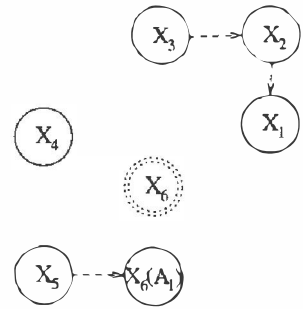

Figure 8: The partial ordering of $EVPI_M(A_1|X_i)$ for the decision model depicted in Figure 7.

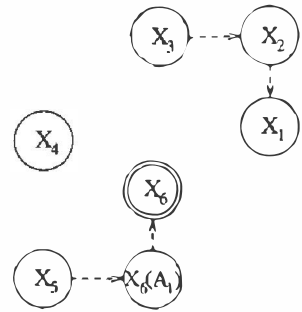

Figure 9: The partial ordering of $EVPI_M(A_2|X_i)$ for the decision model depicted in Figure 7.

**Theorem 5** *Let $M = (C, D, V, E)$ be a decision model, $A \in D$ a decision node, $X \in C$ and $Y \in C$ be distinct chance nodes. If $EVPI_M(A|X) \geq EVPI_M(A|Y)$ and $Cost(X) < Cost(Y)$ then $NEVPI_M(A|X) > NEVPI_M(A|Y)$.*

The result can be used to prioritize information gathering, or to order the attention given by a decision analyst to variables in a decision model drawn from a set of unassessed candidate variables.

## 5 Opportunities for Real-World Applications

The qualitative relationships of the informational relevance of variables in a graphical decision model can be harnessed to make decisions about the relative value of expending effort to acquire information about variables in a decision model, based solely on the topological relationships in the model. Applications of the methods range from obvious, well-understood tasks to longer-range research opportunities.

In a straightforward application, a decision analyst can employ the qualitative analyses during the information-gathering phase of a decision-analysis cycle (Howard, 1966a). For example, assume that an analyst has constructed the decision model depicted in Figure 5 and is considering gathering information to resolve key uncertainties. Which variables should be examined first? In principle, the analyst could compute the EVPI values for all of the variables and choose the variable with the highest expected value of perfect information. However, armed with the results described in this paper, the analyst could employ a topological analysis to eliminate candidate variables. For example, in the model depicted in Figure 5, if $NEVPI_M(A|X_7) < NEVPI_M(A|X_5)$, an analyst can rule out $X_7$ as the next most important variable to focus on. We can repeat this type of analysis to eliminate other candidates.

The results can be similarly used to identify an ordering over the next best test to perform or information to gather in decision-theoretic diagnostic systems such as the Pathfinder system (Heckerman, Horvitz, & Nathwani, 1989; Heckerman, 1991). At any point in a consultative session with such a system, there is an opportunity to gather more information. Quantitative analysis of EVPI for variables can be performed quickly in simple decision problems (Jensen & Liang, 1994). However, evaluating the value of information for large sets of chance nodes and complex sequences of decisions can be computationally costly. In practice, approximation methods have been employed for computing EVPI for large problems. Approximations for EVPI include the popular use of relative entropy and related metrics (Ben-Bassat, 1978; Ben-Bassat & Teeni, 1984; Heckerman et al., 1989; Horvitz, Heckerman, Ng, & Nathwani, 1989a). More recent approximations have explored the use of the statistical properties of large samples to develop value of information approximations (Heckerman, Horvitz, & Middleton, 1991). Our results may be useful in decision-theoretic diagnostic systems for providing an ordering over findings that are most useful for disciminating among hypotheses, with little or no numerical computation at all.

Moving beyond gathering information, the qualitative analysis of EVPI can be used to guide the refinement of decision models. Ordering variables by EVPI can help to prioritize the effort allocated to refining specific variables, definitions, and relationships in a decision model. As we mentioned at the outset of this paper, our work on EVPI was an extension of earlier work on EVR for different dimensions of decision-model refinement (Poh & Horvitz, 1993). The EVR measures are analogs of the value of information. We can employ the relationships developed in this paper to control the sequencing of effort in model refinement.

A particularly promising application of automated control of model refinement is the guidance of the knowledge-based construction of decision models. There has been growing interest in the automated construction of decision models by logical reasoning system (Breese, 1987; Wellman, 1988; Haddawy, Doan, & Goodwin, 1995; Goldman & Charniak, 1990). There is opportunity for combining the qualitative EVPI relationships with work on EVR to automate the control of model construction in applications that employ automated procedures for building and solving decision problems. Such methods hold particular opportunity for such cases as building and solving decision models in time-critical situations (Horvitz, Cooper, & Heckerman, 1989; Horvitz, 1990; Breese & Horvitz, 1990).

The qualitative analysis of informational relevance can also provide a set of handles for controlling and characterizing the error on results generated by inference approximation procedures. The qualitative relationships described in this paper, as well as related results, have been harnessed recently in research on approximate Bayesian-network inference (Liu & Wellman, 1996). Moving beyond probabilistic inference, there is opportunity to develop utility-directed analogs of probabilistic inference algorithms for performing decision-theoretic inference with decision models. Having immediate access to an ordering over the informational relevance of variables in a graphical decision model can be used to control the focus of attention of approximation algorithms, with a goal of minimizing expected cost associated with the approximation. In pursuing utility-directed control, investigators may be able to leverage earlier work on adapting Bayesian network algorithms to decision-theoretic inference in influence diagrams (Cooper, 1988; Peot & Shachter, 1991). Several different classes of Bayesian network inference-approximation algorithms are potential substrates for developing new approximation strategies that might be controlled by qualitative EVPI analyses. These include algorithms that perform approximate inference by simplifying and sequentializing difficult problems via operations on Bayesian networks such as search (Cooper, 1984), conditioning, (Horvitz, Suermondt, & Cooper, 1989b; Dagum & Horvitz, 1992) abstraction (Wellman & Liu, 1994), partial evaluation (Draper & Hanks, 1994), or pruning and clustering (Draper, 1995).

Another area of application of the methods is the guidance of experimentation and learning, given costly information. To date, techniques for learning Bayesian networks and influence diagrams from data focus largely on the case where some static quantity of data is available for analysis (Cooper & Herskovits, 1991; Heckerman, 1995; Buntine, 1995). In real-world learning, we must often consider the costs and benefits of different kinds of data. In the general case, we are forced to make decisions about which data to gather next, given the currently available data set and a decision or set of decisions that must be made. The qualitative relationships about informational value can be employed in decisions about the most critical data to gather, and about the most important model structures and potential hidden variables to search over.

## 6  Summary and Conclusions

We have developed a qualitative analysis of EVPI in influence diagrams. The methods provide an ordering over EVPI values for variables in an influence diagram based on the topological relationships among variables in the model. We described a procedure for identifying a partial order over variables in terms of their EVPI. The resulting partial ordering can be represented by a graph which is a subgraph of the original influence diagram.

Similar to other qualitative analyses, the method does not always provide an EVPI ordering for variables. To resolve important ambiguities about EVPI or NEVPI, we can employ a targeted quantitative analyses that considers the details of the underlying utility model and probability distributions. Nevertheless, even when our qualitative analysis fails to produce a total ordering, we can obtain a partial order on the EVPI of variables via the procedure we described. Moreover, if one or more of the actual EVPI values are known, the qualitative analysis can provide upper or lower bounds for EVPI with only minimal computational effort. Such information can be valuable to a decision analyst for directing attention to information that would yield maximal returns and to identify the topological position of most important, but as yet unmodeled, conditioning events. The methods also show promise for providing automated reasoning systems with efficient mechanisms for determining the most important variables and information to focus attention on during model refinement or decision-making inference.

## Acknowledgments

We thank Chaolin Liu and Michael Wellman for comments on an earlier version of this paper. We are also grateful for the valuable feedback provided by the anonymous referees.